# Towards a Generic Multimodal Architecture for Batch and Streaming Big Data Integration

[1,3]Siham Yousfi, [2,3]Maryem Rhanoui and [1]Dalila Chiadmi

[1]*SIP Research Team, Rabat IT Center, EMI, Mohammed V University in Rabat, Morocco*
[2]*IMS Team, ADMIR Laboratory, Rabat IT Center, ENSIAS, Mohammed V University in Rabat, Morocco*
[3]*Meridian Team, LYRICA Laboratory, School of Information Sciences, Rabat, Morocco*

**Abstract:** Big Data are rapidly produced from various heterogeneous data sources. They are of different types (text, image, video or audio) and have different levels of reliability and completeness. One of the most interesting architectures that deal with the large amount of emerging data at high velocity is called the lambda architecture. In fact, it combines two different processing layers namely batch and speed layers, each providing specific views of data while ensuring robustness, fast and scalable data processing. However, most papers dealing with the lambda architecture are focusing one single type of data generally produced by a single data source. Besides, the layers of the architecture are implemented independently, or, at best, are combined to perform basic processing without assessing either the data reliability or completeness. Therefore, inspired by the lambda architecture, we propose in this paper a generic multimodal architecture that combines both batch and streaming processing in order to build a complete, global and accurate insight in near-real-time based on the knowledge extracted from multiple heterogeneous Big Data sources. Our architecture uses batch processing to analyze the data structures and contents, build the learning models and calculate the reliability index of the involved sources, while the streaming processing uses the built-in models of the batch layer to immediately process incoming data and rapidly provide results. We validate our architecture in the context of urban traffic management systems in order to detect congestions.

**Keywords:** Big Data Integration, Lambda Architecture, Heterogeneous Data, Urban Traffic Management Systems

## Introduction

Big data are a set of large scale heterogeneous data that flew at high velocity. They are produced from a variety of channels including social media, crowdsourcing platforms and sensors. Big Data mining helps companies in various sectors such as smart cities, healthcare, e-learning etc. to analyze their activities, understand the users' actions and predict their behaviors. Nevertheless, Big Data sources can have different levels of heterogeneity; they can provide data with different types (text, image, video or audio), formats (xml, json, csv, jpg, avi, etc.) and levels of accuracy and completeness. Therefore, in order to build a global accurate and complete view over Big Data, it is important to adopt a multimodal architecture that can extract the knowledge from all available types and formats of data.

Offline Big data analysis is very useful to extract insight, provide models for statistical analysis and correct predictions. However, despite being accurate, the insight value is decreasing overtime and should be completed by the real-time analysis. In fact, in some cases it is crucial to make decisions based on events as they occur. This requires providing the right data at the right moment. To combine historical and streaming data analysis, the Lambda Architecture is a paradigm that has been widely adopted in both industry and academia. This paradigm is designed to process large volume of historical data, as well as rapid incoming data streams. It is composed by three layers; batch layer, speed layer and serving layer. The first two layers are complementary since each one has its own advantages and limitations. From one hand, the batch layer allows deep processing of the whole dataset which provides accurate views of data. However, the processing time is slow and may take hours. From the other hand, the speed layer allows creating stream views by immediately processing incoming data. However, the results may lack of accuracy. The results are finally merged in the serving Layer that provides an interface for queries.

In the present paper, we propose a generic multimodal architecture inspired by the lambda architecture to handle both batch and streaming data processing. Our architecture is not limited to a simple instantiation of Lambda Architecture. It also allows the streaming layer to reuse the updated built-in models and knowledge bases during the batch processing. Moreover, our architecture deals with multiple heterogeneous data sources and aims to ensure data reliability and completeness of the final insight.

Our solution is generic and could be applied to many contexts that deal with heterogeneous data sources that provide large volume of data at high velocity and require both batch and streaming data analysis. For example, in health care domain, the historical patient's data provided by medical sensors are useful to build models and predict the evolution of patient's conditions when the critical heart or blood pressure alerts should be processed once generated in order to prevent potential immediate death. Another example concerns smart urban transportation where traffic data are produced by multiple heterogeneous data sources including GPS, loop detectors, CCTV and aerial vehicles. Historical traffic can help city managers identify roads states, predicting congestions and making good decisions about roads planning and development. While the incoming streams of traffic data can be used to provide citizens with instant insight and alerts about the real-time traffic state.

We choose the second example to validate our architecture. In fact, we implement our solution using several Big Data tools and libraries mainly based on spark environment since it offers optimal implementation of architectures combining both batch and streaming processing modes. For the batch layer, we use spark for Natural Language Processing and log processing, OpenCV to analyze images data and Matlab for videos processing. Regarding the speed layer, we use apache Kafka for message queuing along with spark streaming for stream processing.

The remainder of this paper is organized as follow; the first section presents the background of the lambda architecture and the second section summarizes the related works according to two main topics; Multimodal Architecture for Big Data Integration and Lambda Architecture implementations. In the third section, we describe our multimodal architecture for streaming Big Data integration. Our architecture is finally validated in the context of urban traffic management systems in the last section.

## Background: Lambda Architecture

The Lambda architecture is a novel concept introduced by Marz and Warren (2015) in order to handle large scale data and to solve the problem of real-time processing. The Lambda Architecture aims to meet the needs of a robust, fault-tolerant system, to serve a wide range of use cases in which low latency reads and writes are required. The resulting system must be linearly scalable.

The Lambda Architecture is composed of three layers: the batch layer, the speed layer and the serving layer each, performing specific functionalities:

- The batch layer stores a copy of a very large dataset and precomputes batch views. Batch-processing dataset and compute random functions on it

  batch view = function(all data)

- The serving layer is a distributed database that supports batch updates and random reads. It loads in a batch view and enable to do random reads on it.g)
- The speed layer aims to provide efficiently queried views that contain recent data. The real-time views are updated when receiving the new data which are combined with the previously computed real-time views

  realtime view = function(realtime view, new data)

- The final query is applied on both views:

  query = function(batch view. realtime view)

The lambda architecture has the advantage of being technology independent and disregarding infrastructure. It provides users with the freshest possible data views along with a scalable historical data. Nevertheless, as shown in Fig. 1 the initial architecture does not consider communication between batch and speed layers while it would be interesting to use, in speed layer, the models previously trained in batch layer. This will avoid too much processing and reduce response time (Baldominos *et al*., 2014). Moreover, the lambda architecture focuses on the processing mode and doesn't consider the heterogeneity of data sources which can have a negative impact on the accuracy and completeness of the final views. Indeed, batch and streaming modes can provide inconsistent results. This is because the batch views can be of a high level of accuracy since processing large amount of data allows performing comparison and matching while requiring a long processing time. Likewise, the streaming mode, although faster, provides information with a low level of accuracy. Therefore, the user's queries should contain additional processing in order to evaluate the data accuracy and merge the results coming from both layers.

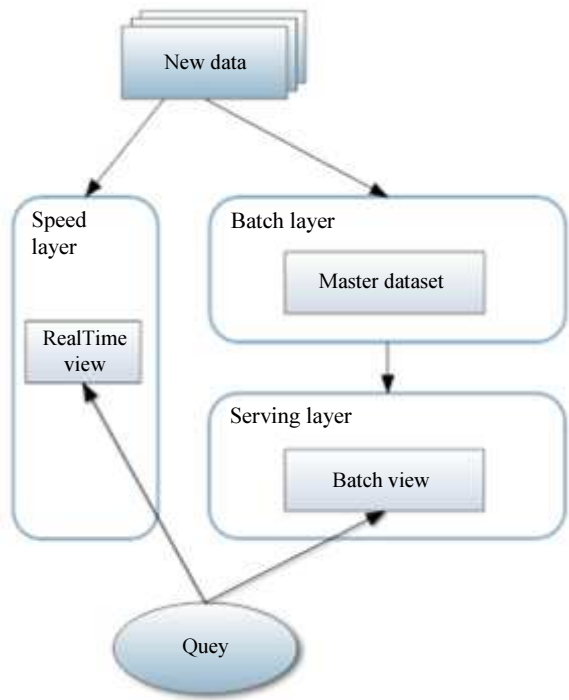

**Fig. 1:** Lambda Architecture (Marz and Warren, 2015)

# Related Works

There are two main topics for the related works. The first one deals with the multimodal architecture for Big Data Integration and the second one focuses on the implementation and integration of both batch and streaming modes of the Lambda architecture. We describe and discuss bellow the related works for each topic.

*Multimodal Architecture for Big Data Integration*

Klein *et al*. (2016) proposes a reference architecture for Big Data applied to the national security domain. The architecture decomposes the system into 13 modules divided into Big Data Application Provider modules, Big Data Framework Provider modules and Cross-Cutting modules. Within the Big Data application provider module, Data Processing and Integration Module transforms data and transfers it to the other modules using the traditional Extract, Transform Load (ETL) method. The transformed data is analyzed by the Data Analysis Module in order to extract relevant information. In addition, Chen *et al*. (2016) proposes a multi-source and heterogeneous Big Data integration model. They use an improved ETL framework named BDETL to integrate the data form heterogeneous data sources in a distributed environment. The model has been implemented in a power dispatching and control system. However, the heterogeneous data sources consist mainly on different operational systems data and thus do not handle different data types. Costa and Santos (2016) proposed BASIS, a Big Data architecture applied to Smart Cities based on the separation of abstraction layers, namely the conceptual, the technological and the infrastructural layers. Finally, Amini *et al*. (2017) proposed Big Data Analytics architecture based on distributed computing platform for real-time traffic control. The architecture is flexible and modular as it handles multiple data sources. A partial prototype has been developed using Kafka and python for the data analysis.

As a conclusion, the related work for this topic are either limited to the analysis of a single data source or propose a unimodal architecture dealing with multiple data sources but providing the same data type or at best suggest to analyze multiple data sources with no working practical validation case presenting and explaining the multimodality. Moreover, the considered data sources are not distinguished according to their reliability or the completeness of the data they provide.

*Implementation of Lambda Architecture*

Jambi and Anderson (2016) present an implementation of the lambda architecture for real-time crisis events exploration. The reference presents clearly the system components as long as the way they interact with each other. The authors used Apache Kafka for distributed data messaging and queueing, spark for batch and stream processing and Apache Cassandra for event and twitter data storage. However, the analysis is limited to metadata and hashtags and doesn't deal with the content of the tweet. Moreover, Khazaei *et al*. (2016) present the Sipresk Big Data platform as an implementation of the lambda architecture using spark environment. The platform analyzes loop detectors data and is able to classify traffic events to short, medium, long or extended according to their duration. The results are displayed in interactive map with timestamp.

In addition, Hasani *et al*. (2014) defend the use of the lambda architecture in order to perform real-time Big Data analysis. However, they implement only the batch layer using the Hadoop framework and don't propose a complete and working implementation of the real-time layer. Twardowski and Ryzko (2014) propose a lambda-based architecture for Big Data processing using multi-agent systems. The reference illustrates how autonomous agents can be used in order to exchange data between the processing layers of the lambda architecture and provide capabilities for robust processing of data in real-time. A theoretical implementation for collaborative filtering recommender system was presented. Baldominos *et al*. (2014) propose a machine learning architecture for Big Data based on lambda architecture allowing batch and stream analysis. The architecture is composed by four elementary modules: The batch machine-learning module that performs query analysis, data clustering and machine learning models building, the stream processing module that takes advantage of machine learning models previously built by the batch processor in order to make classifications and predictions and provide recommendation, the storage module and the dashboard module. Baldominos *et al.* (2014) developed two systems for validation: A recommender system for web advertising and a prediction system for gamers' behaviors. Finally, Kiran *et al.* (2015) implement the Lambda Architecture design pattern to handle sensors and smart phones data in order to optimize network costs. It provides affordable and cost effective real-time Big Data processing that can be applied in any scenario by combining database management, query management and cloud computing. They applied their solution by processing router sensor data on the ESnet network.

The related works for this topic are either limited to a single layer (batch or streaming), a single data type or perform basic processing on both layers without merging the extracted knowledge.

## Discussion

After going through the different related works suggesting either an architecture for Big Data integration or an implementation of the lambda architecture, we can conclude that the proposed working architectures, platforms or solutions are those that reduce the complexity of the Big Data environment by focusing on a single data source, a single data type or a single processing type (batch or a streaming). However, the analysis results are likely to be incomplete (when some data sources are neglected), inaccurate (when the analyzed data sources are unreliable) or outdated (when data are processed a long time after being produced). Therefore, in this paper we propose a generic architecture that provides a near-real time global and accurate view of Big Data collected from multiple and heterogeneous sources. For that purpose, our architecture combines batch and streaming multimodal processing for all the available data types and formats to ensure maximum completeness of results and compares data coming from several sources to evaluate their reliability.

## Generic Multimodal Architecture for Streaming Big Data Integration (Streaming-Ma4BDI)

The Streaming-Ma4BDI architecture is inspired by the lambda architecture. As shown in Fig. 2 the newly received data streams are routed to both batch and streaming processing modes. The first one stores the data stream in a distributed storage media waiting for the next batch iteration to be processed. It performs offline processing tasks such as multimodal data analysis, building learning models, calculating and updating sources reliability index which is assigned to each source and indicates its reliability level. The second is the streaming processing mode which processes data at reception and provides real time or near-real time views. This mode uses the reliability index as well as the learning models already computed during batch processing mode. The last component of our architecture is the query engine which offers to users a global view based on both stream views and batch views.

The implementation of the streaming mode allows handling data as soon as they are received. It complements the batch mode. We detail below the different modes and layers of our architecture and discuss the differences between them.

### Batch Processing Mode

The batch processing creates views incrementally. It collects data from various Big Data sources and performs multimodal processing through various engines each one intended for a given data type: Text, image, video and audio. These engines use specific learning models to extract metadata (mi) and knowledge (ki) from the stored data. The knowledge is then merged in order to provide the most accurate and complete insight based on sources reliability index. The batch layer is intended for all periodic non-time-critical tasks such as building and updating learning models, reliability index computation and knowledge base updating etc.

Figure 3 presents a high level overview of the batch mode. We describe in detail in the following each of its components:

### Big Data Sources Layer

The Big Data sources (si) provide data with different levels of heterogeneity including:

- Data type: The data sources can generate structured, semi-structured and unstructured data including text, images and videos
- Data format: For the same data type, several data formats can be used. For example textual data may be provided in XML, JSON, TXT, etc
- Data accuracy: That depends on the reliability of the sources data are extracted from
- Data coverage (completeness) since some collected data can be more complete than others
- Extracted knowledge; the data may concern different events and even for those concerning the same event they may provide contradictory or complementary information

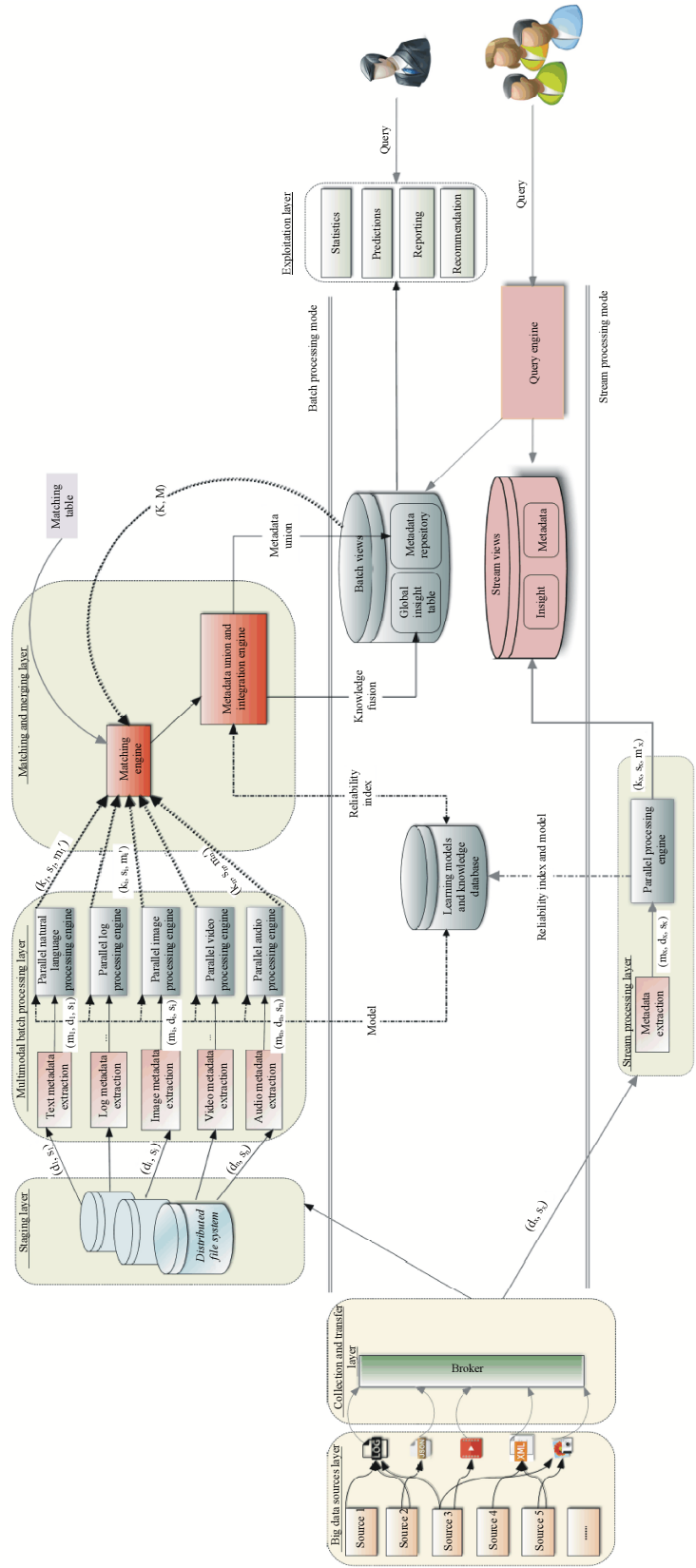

**Fig. 2:** Generic multimodal architecture for streaming big data integration

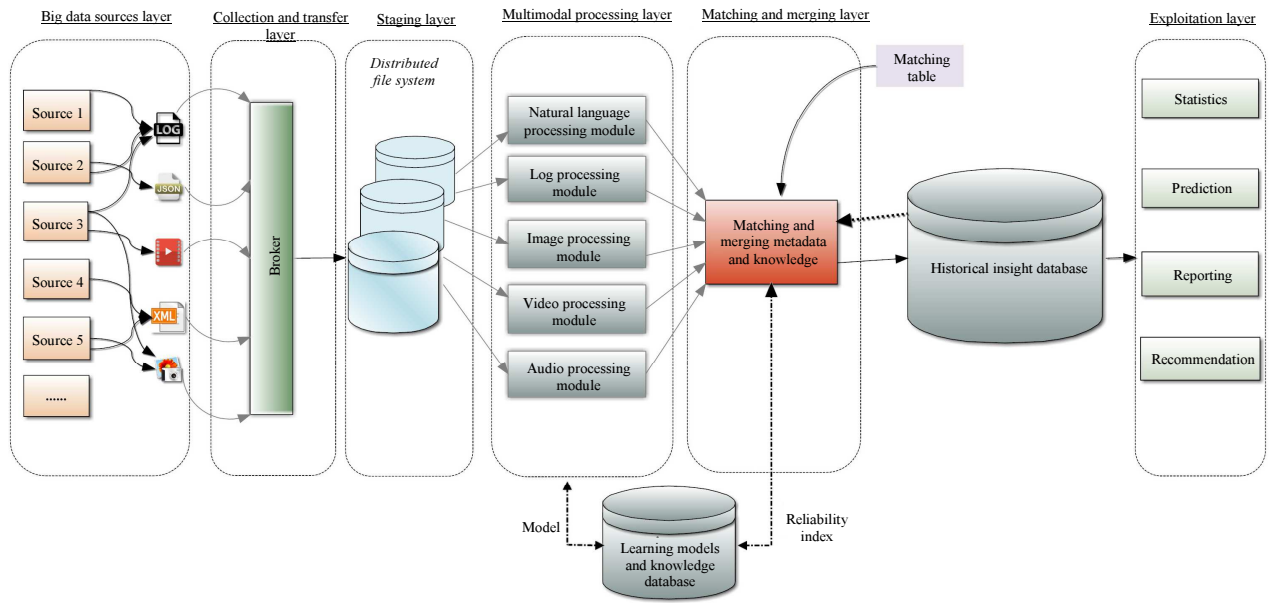

**Fig. 3:** Overview of the batch mode of the Streaming-Ma4BDI

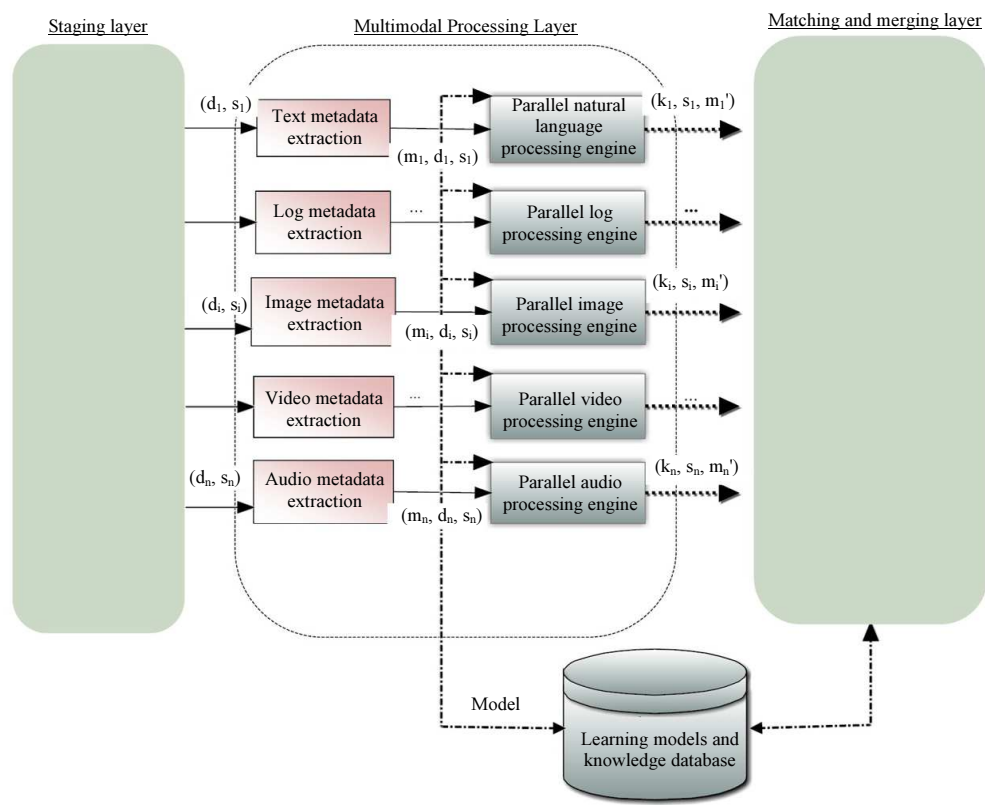

**Fig. 4:** Multimodal processing layer

Collection and Transfer Layer: the collection and transfer layer collects data related to a specific context, detects the data type of each dataset and routes them to the corresponding storage space.

Staging Layer: The staging layer stores data automatically in a distributed storage system and separates the different data according to their type (text, image, video and audio).

Multimodal Batch Processing layer: As shown in Fig. 4, the processing layer consists of two main tasks; Metadata extraction engines and data processing engines. Multimodality is provided in this layer through the implementation of a specific engine for each type of data:

- Metadata extraction engine: Receives a pair of ($d_i$, $s_i$) and sends the triple ($d_i$, $s_i$, $m_i$) where

    - $d_i$ is the raw data
    - $s_i$ is a set of information about the data source providing di, such as name of author and URL
    - $m_i$ is the available relevant metadata extracted from the file's description

- Data processing engine: It contains several processing engines each using a specific mode. They aim to clean, normalize and complete metadata before extracting relevant. Also, this engine can perform further processing in order to update or complete the metadata $m_i$. It provides a triple ($k_i$, $s_i$, $m_{i'}$) where:

    - $k_i$ is the extracted knowledge
    - $m_{i'}$ is the updated version of metadata mi

The data processing engines use pre-built learning models such as machine learning or deep learning models in order to extract knowledge $k_i$. In addition, these engines could be used to update the knowledge database and refine the learning models

Matching and merging layer: The matching and merging layer discovers redundant, complementary and conflicting knowledge sent by the Multimodal Processing engines. As shown in Fig. 5, this layer uses the matching engine in order to compare the sets of ($k_i$, $s_i$, $m_{i'}$) triples resulting from the processing layer and the global knowledge ($K$, $M$) already stored in the historical insight database (if exists) (1). The matching engine identifies records concerning the same event and eliminates redundancies.

The Metadata union and Knowledge Merging engine performs a union of metadata $m_i$ in $M'$ and merges knowledge $k_i$ in $K'$ (2). In addition, it calculates and updates the reliability index of each data source (3) and decides which to trust in case of conflicts. Finally, it stores the global insight represented by the pair ($K'$, $M'$) in the historical insight database (4), (5).

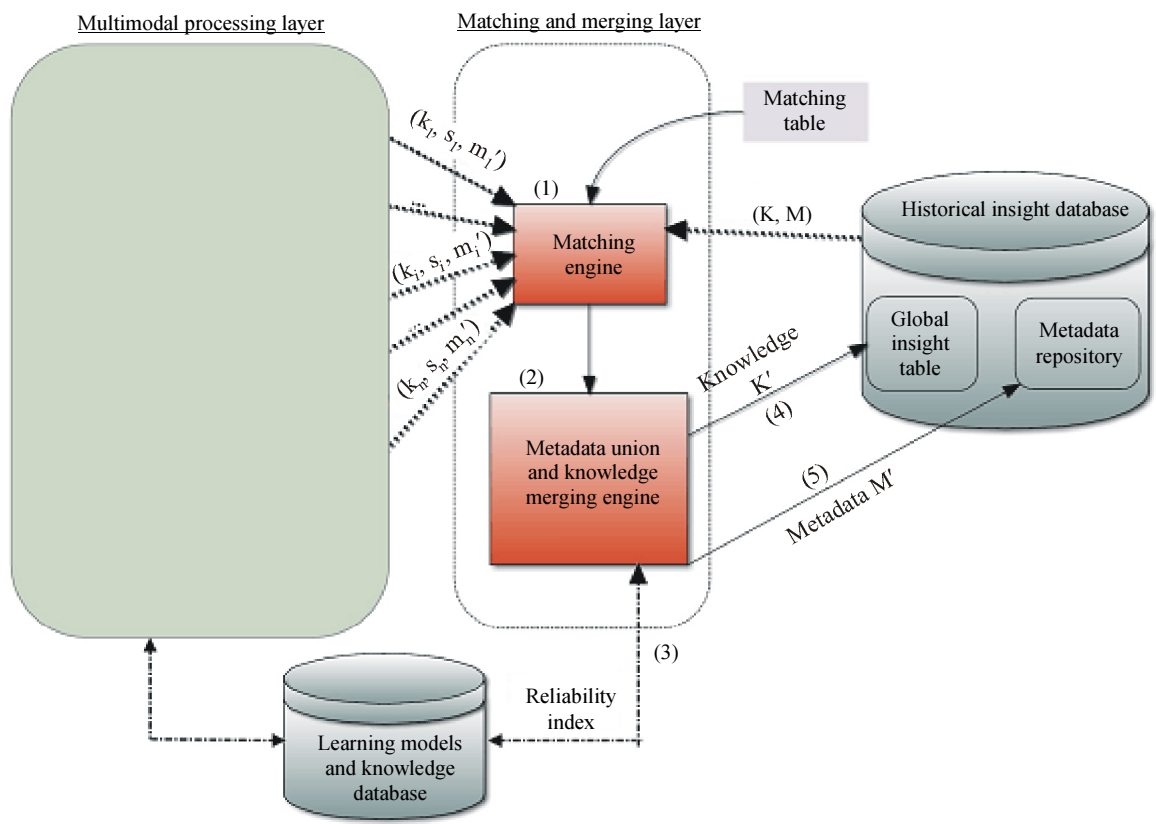

**Fig. 5:** Matching and merging layer

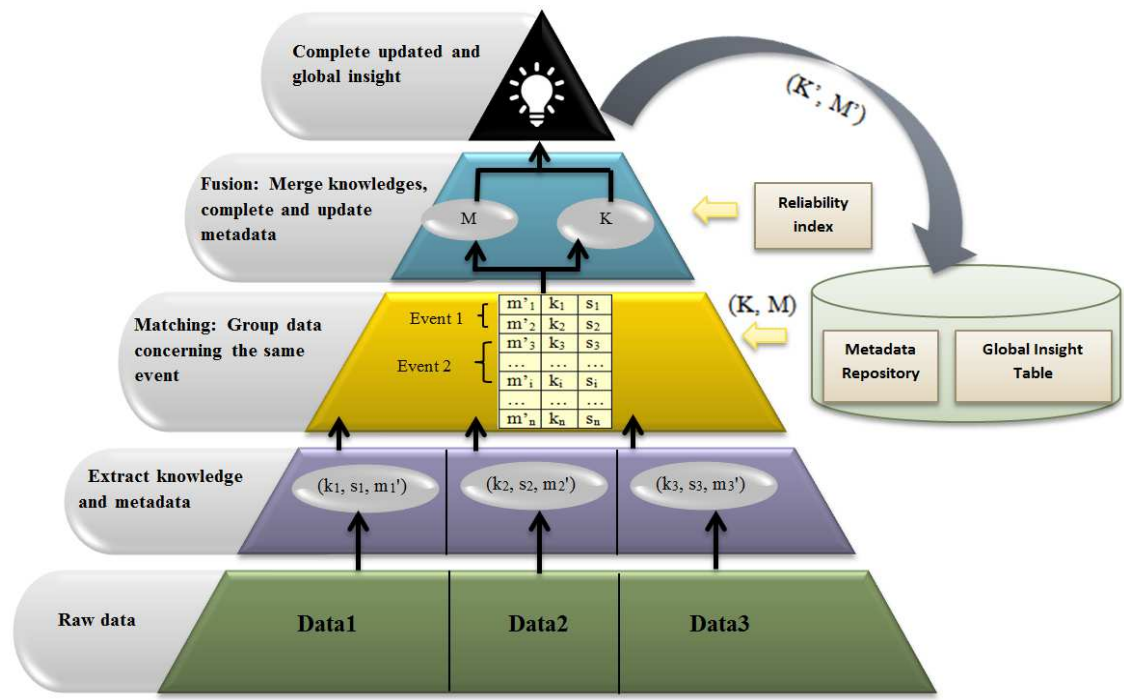

**Fig. 6:** Global Insight construction

Figure 6 shows the overall process to extract knowledge and construct the global insight. The figure describes an example with three datasets $d_1$, $d_2$ and $d_3$ gathered from multiple data sources with $d_1$ and $d_2$ concerning the same event. Those data provide respectively the triplets $(k_1, s_1, m_{1'})$, $(k_2, s_2, m_{2'})$ and $(k_3, s_3, m_{3'})$ after processing. Before final storage, it is necessary to know if the events these triplets describe has already been identified and stored in the historical insight database. This step is carried out through a matching between each triplet and the old global insights represented by the set of (K, M) pairs. Then a fusion step is performed by the Metadata Union and Knowledge Merging engine. Indeed, this engine does double actions: (1) Union of metadata in M' to best approaching data completeness and (2) fusion of knowledge based on the reliability index of each data source in K'. The pair (K', M') constitutes the final global insight. It is computed using fusing algorithms such as majority votes. The reliability index of each data source is updated according to results. In fact, it is increased for data sources that made the right prediction and decreased for those who made the wrong one.

Exploitation layer: The exploitation layer allows users to access the insights stored in the final database through statistics, predictions, recommendations etc.

The objective of the batch mode of our architecture is to analyze data gathered from multiple Big Data sources and integrate the corresponding knowledge in order to provide most complete and accurate global insight. The whole operation is a parallel large scale process that is performed iteratively. Though, depending on the data size, each iteration may take a long time to be executed. Therefore, the batch process results are always outdated since new data are received during batch execution.

The batch processing mode is intended for historical data analysis, the high latency computation of the system is not considered as a limitation since these data will clearly be processed during the next batch. However, in some cases, the value of data is decreasing over time and extracting corresponding knowledge once generated is much useful. This leads to a high need for building views on data and computing knowledge at reception. In order to overcome this limitation we use the streaming mode is proposed.

*Stream Processing Mode*

The streaming processing mode aims to immediately process incoming data. This mode concerns the data sources providing data at high velocity. The broker identifies the data type and starts up the streaming process by routing the data to the appropriate engine according to their type. In order to lighten the processing and meet real-time and near-real-time needs, we chose to avoid I/O tasks at this level and transfer data without intermediate storage. In the streaming mode, the processing layer allows the extraction of the knowledge as well as the corresponding metadata according to the type of data as explained below:

The Metadata extraction engine input is a flow of data-pairs ($d_i$, $s_i$). The Data processing Layer uses one of the five processing engines of the Multimodal Batch Processing Layer in order to build a cleaned, normalized and completed metadata version $m_{i'}$ and extract the knowledge $k_i$ based on the learning models already created during the batch mode. Therefore, data processing engines provide the triples ($k_i$, $s_i$, $m_{i'}$) which are stored in the stream views. Note that the built and refinement of each learning model is done exclusively by the batch layer resulting in a significant reduction of the processing time of the stream processing layer.

The streaming mode continually updates "stream views" with the quadruplet ($k_i$, $s_i$, $m_{i'}$, $r_i$) where $r_i$ is the reliability index of the knowledge $k_i$ calculated during the batch processing mode. This index is based on several parameters that identify the reputation of the data source such as its identity, author, the rate of correct information that it produced in the past etc.

Both batch and streaming layers complement each other. Indeed, although the batch processing does not satisfy the temporal constraints, the insight it provides has a great value given its high level of reliability. The latest (insight) is obtained from knowledge comparison performed continually to update the Big Data sources reliability index. Instead, the streaming mode that manages each data at reception does not take advantage of the huge mass of input data to compare and match knowledge, which may impact the accuracy of results. Nevertheless, since the accuracy of knowledge depends on the reliability of the data source it comes from which is already calculated in batch mode, the streaming layer uses this index to identify the accuracy of knowledge.

The batch and streaming modes share common processing steps such as metadata extraction and knowledge discovering. We could think about capitalizing these common steps in the streaming layer and make the batch layer using the results in order to avoid redundancy. However, this requires adding, in the streaming layer, a supplementary I/O tasks to store the extracted metadata and knowledge so as to be used later by the batch layer. This I/O task will slow down the streaming process and we believe that it is better to repeat the same process in the batch layer since no temporal constraints are imposed.

*Query Engine*

The query layer allows random access to the insights in real-time or near-real-time. Indeed, at the user request, the query layer will first look for relevant insight in the "stream views". In case this information is not available the request is transmitted to the "batch views" where previously built models will be used in order to find or predict the insight. The final result can be available to users through graph, dashboard etc.

The architecture we present is generic and may be applied in any context that deals with heterogeneous data sources which provide huge volume of data at high velocity and require the combination of batch and streaming data analysis. The following section aims to validate our solution through a real example of urban traffic management systems. It also highlights some specific technical and functional details that belong to the implementation of our architecture.

## Case of Study: Urban Traffic Management System

We apply our architecture to the context of smart city, especially the urban smart mobility to analyze traffic data in order to identify congestions. Our solution provides city managers and citizens with global accurate views about traffic status in near-real-time which helps analyze and improve the quality of urban transportation and recommending alternative less congested roads to citizens.

More than 54% of the world's population was living in urban areas in 2016 and has reached 66% by 2030 (?), (?. The cities are growing rapidly and the need for vehicles is increasing, the low availability and the long delays of public transportation led households to attempt to own at least one tourist car which leads to permanent congestion, particularly in big cities, resulting in a negative impact on economy, society and environment.

To reduce the urban traffic congestion we propose a Big Data solution that provides cities managers with complete and accurate insights about the urban traffic status. Our solution analyzes traffic data coming from heterogeneous sources to detect traffic events. The later can be an accident, a slow traffic or a number of vehicles exceeding the capacity of the road, construction or maintenance work, sportive or social events etc. Our solution allows managers to analyze the causes of daily urban congestions, predict future congestions and therefore take the required actions for their debottlenecking such as the widening of the roads, the creation of new sections etc.

*Data Sources Layer*

The data sources involved in urban traffic context are sensors, CCTVs, aerial vehicles, social Medias, electronic newspapers and crowdsourcing applications. None of these data sources provides complete and accurate information related to one road under varied conditions (Weather, day/night etc.). Moreover, besides the heterogeneity of the data formats and types, the knowledge extracted from these sources is also heterogeneous since it could be complementary, redundant or contradictory. It is necessary to integrate knowledge gathered from these multiple data sources in order to build a complete, global and accurate insight about traffic state.

For the validation purpose of our solution we are using real datasets that we apply for a city named "XCity". In the batch mode of the present paper we are considering all the data sources presented in Table 1 as follow:

- Social media: Tweets concerning traffic events (accidents, traffic jams, etc.)
- Online newspaper: Web pages from many electronic newspapers such as: CBS Chicago, Chicago tribune etc
- Sensors: Traffic data collected from Loop detectors
- Geo-location data: Traffic GPS data including the latitude, longitude, elevation, date and time
- Unmanned Aerial vehicle: Traffic image data
- Closed-circuit television: Traffic video data

However, in the streaming mode, we are using Twitter only as this data source has largely been used by citizens and administrations for publishing traffic incidents. In fact, Ribeiro *et al*. (2012) propose a system to detect and locate traffic events with Twitter in Belo Horizonte. They found that there is a significant correlation between real traffic conditions and tweets talking about traffic conditions. Moreover, Tian *et al*. (2016) has evaluated the quality of traffic event tweets in Austin, Texas. The study has proved that citizens tweet more often in case of true and major severe incidents compared to false and minor incidents. Finally Rashidi *et al*. (2017) stated that social media can be considered as a supplementary source of data to extract complementary information about traffic conditions.

*Implementation of the Streaming and Multimodal Big Data Integration solution*

For validation purpose, we discuss in this section the implementation of our architecture that aims to handle historical and real-time coming data. We describe in this section the technical architecture of our solution as well as final results.

*Technical Architecture of Streaming-MA4BDI*

Figure 7 shows the technical architecture of our solution based on Apache Spark environment in order to ensure interoperability and code reuse.

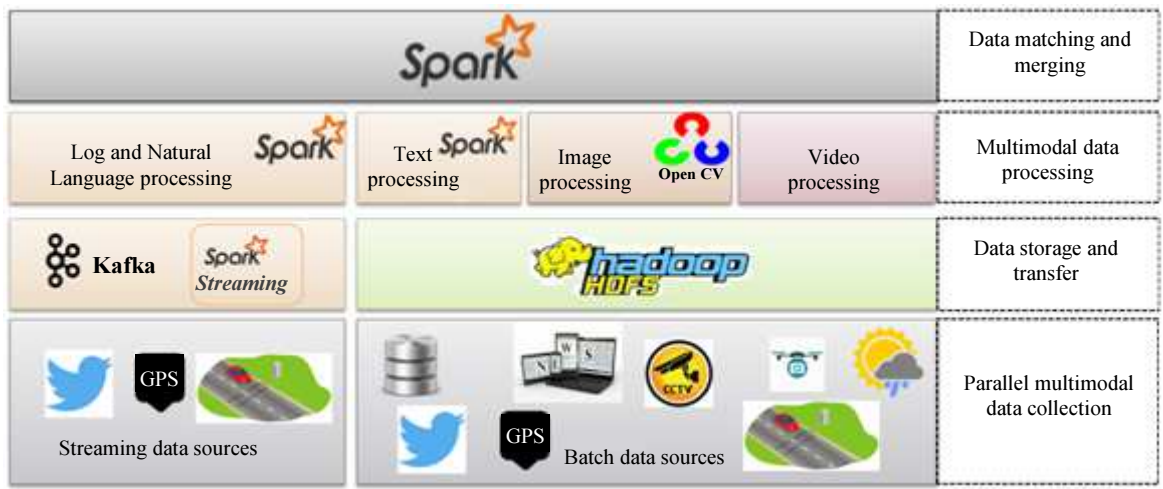

**Fig. 7:** Technical implementation

**Table 1:** Traffic big data sources

| Data source | Information | Data type | Metadata |
|---|---|---|---|
| Sensors | Number of vehicles, traffic speed | Semi-structured text data | Latitude, longitude, time, speed |
| Social media | Congested road, slow traffic, accident, road maintaining and event. | Unstructured Text data | Text, date, time, user.name |
| Online newspapers | Accident, road maintaining and event. | Unstructured Text data | Title, date, time, writer, journal name, content |
| Geolocation data | Traffic speed, number of vehicles | Semi structured text data | Vehicle, latitude, longitude, speed, date, time |
| Unmanned Aerial vehicles | Number of vehicles | Unstructured images | Latitude, longitude, image |
| Relational databases | Official sport, music events, static information about roads. This data source is also used to fuse results | Structured text data | Road name, Id, latitude, longitude |
| Closed-circuit television | Number of vehicles on a special road | Video stream | Latitude, longitude, time, video |
| Weather web service | Weather conditions (will help predict data accuracy) | Structured text data | Date, time, condition (raining, fog, etc.) |

Spark helps implemente the lambda architecture since it handles both batch and streaming analysis:

- The batch processing manages all traffic data sources including CCTV cameras, social media, GPS, loops detectors, UAV and web pages. It uses the Hadoop Distributed File System HDFS to ensure scalable distributed storage and Spark, Open Cv and Matlab to build and used learning models for metadata and knowledge extraction. Results are matched in order to detect data concerning the same event then merged considering the weight of data sources, in order to provide the final global complete and accurate insight
- The streaming processing handles only data sources that provide stream text traffic data. in our case we are analyzing tweets. It uses Apache Kafka, for distributed data messaging and queuing to create data streams and Apache Spark Streaming for parallel natural language processing and model exploitation in order to identify traffic event (knowledge) and extract and complete metadata based on machine learning models built-in during batch mode. Finally, the data sources reliability indices calculated in the batch layer are used in order to identify the accuracy of the extracted knowledge and then provide a accurate global insight about the traffic event
- The query engine uses Apache Spark in order to understand the user's query, look for results in the stream views of Hbase. If found the results are displayed for the user. Otherwise, the layer uses the batch models in order to predict results

*Multimodal Batch Processing Mode*

The batch processing mode analyzes the extracted data from the different traffic sources and stores them separately according to their types (image, video, text) in Hadoop distributed file system. Then, during the first analysis step, the corresponding metadata is extracted such as the date and time of the congestion event, its location and the date and time of resolution (Time to return to normal traffic).

The second analysis step uses multimodal processing engines according to the data type in order to extract the corresponding knowledge namely, congested or not congested. These engines build and exploit machine learning models to analyze tweets text and detect messages related to traffic congestion (Bayesian model), calculate traffic speed from GPS data and loop detectors, analyze the images (support vector machine model) and videos (Gaussian mixture model) provided by CCTVs and aerial vehicles to detect and calculate the number of vehicles on a road and thus identify the traffic state. Table 2 summarizes an example of the knowledge and metadata returned by the processing layer.

The parallel processing engines send the knowledge to the merging layer that matches, compares then merges the extracted knowledge based on a weighting system that assigns to each source a reliability index in [0, 1]. In traffic context, the data source reliability index value is dynamic because it is related to many parameters including hardware, weather, personal and historic of reliability. For example the quality of images provided by a CCTV or an UAV depends on the hardware configuration and may decline under specific weather conditions such as rain and fog. We have associated an initial weight value to each data source. Table 3 presents an example of the used data sources and corresponding Reliability index. UserA represents an official twitter account such as police or urban management accounts. These accounts are supposed reliable and have higher reliability index value than UserB which represents a standard account. Table 3 shows also that the reliability index of some devices can change according to special conditions. For example, ID1 is a CCTV device it is reliability index decreases in case of rain, snow and fog.

We used the triplet (road identifier, date of event, time of event) as a key to match processing layer results in order to identify the data belonging to the same event. Then, the Metadata union and Knowledge Merging engine merges the metadata and knowledge using the majority vote algorithm (Equation (1)). Finally it updates the dynamic weight of the data sources:

$$H(x) = \sum_{i=0}^{n} q_i h_i(x) \qquad (1)$$

where, $q_i$ is the weight associated to each data source and $h_i(X)$ is the knowledge provided by the processing engines.

The process of the dynamic reliability index update aims to increase the reliability index of the data sources that made the right prediction and decrease the data sources reliability index that made the wrong prediction while remaining within an interval between [0, 1]. For example; if the result (road status) of $H(x)$ is congested, then the weight of all the data sources that predicted the status "congested" will increase by 0.05 and the one of the data sources that made the prediction "not congested" will decrease by 0.05.

Using Table 2 and Table 3, we calculate using the majority vote algorithm, the probability that road 100 could be congested or not congested. According to Table 4, we conclude that the probability that a road is congested (0.6) is greater than the probability that a road is not congested (0.25). We can conclude that road 100 is congested (highest probability).

Table 5 presents the final metadata stored in the repository as a union of the matched correct data. The reason of congestion, date and time of resolution have been merged from UserB and UserD data sources.

**Table 2:** Description processing layer results

| Source | Type | Knowledge | Condition | Metadata | | | | | | |
| --- | --- | --- | --- | --- | --- | --- | --- | --- | --- | --- |
| | | | | Road Id | Road Name | Date of event | Time of event | Reason of congestion | Date of resolution | Time of resolution |
| UserA | Official twitter account | Congested | | 100 | Alpha | 11/07/2017 | 08:10 | accident | null | null |
| UserB | Standard twitter account | Congested | | 100 | Alpha | 11/07/2017 | 08:10 | accident | null | null |
| UserC | Crowd sourcing application | Not congested | | 100 | Alpha | 11/07/2017 | 08:10 | null | null | null |
| UserD | Web newspapers | Congested | | 100 | Alpha | 11/07/2017 | 08:18 | null | 11/07/2017 | 18:00 |
| ID1 | CCTV | Not congested | Rain | 100 | Alpha | 11/07/2017 | 08:15 | null | null | null |

**Table 3:** Description of some traffic Big Data sources reliability index

| Data source's ID | Type | Condition | Reliability index |
| --- | --- | --- | --- |
| UserA | Official twitter account | | 0.30 |
| UserB | Standard twitter account | | 0.15 |
| UserC | Crowdsourcing platform | | 0.15 |
| UserD | Web newspapers | | 0.15 |
| ID1 | CCTV | Rain, snow, fog | 0.10 |
| ID1 | CCTV | Clear | 0.15 |

**Table 4:** Probability values of congestion of road 100

| Status | Probability |
| --- | --- |
| Congested | 0.6 |
| Not congested | 0.25 |

**Table 5:** final complete and merged metadata

| Road Id | Road name | Date of event | Time of event | Reason of congestion | Date of resolution | Time of resolution |
| --- | --- | --- | --- | --- | --- | --- |
| 100 | Alpha | 01/02/2017 | 08:18 | Accident | 01/02/2017 | 10:00 |

**Table 6:** Updated values of Big Data sources reliability index

| Data source | Type | Condition | Reliability index |
| --- | --- | --- | --- |
| UserA | Official twitter account | | 0.30 |
| UserB | Standard twitter account | | 0.20 |
| UserC | Crowd sourcing platform | | 0.10 |
| UserD | Web newspapers | | 0.20 |
| ID1 | CCTV | Rain, snow, fog | 0.05 |
| ID1 | CCTV | Clear | 0.15 |

Finally, the batch processing updates the reliability index for each data sources in order to decrease the weight of the data sources that have predicted the road 100 as not congested (UserC and ID1) and increase the weight for data sources that have predicted the road 100 as congested (UserB and UserD). The couple (metadata, knowledge) is stored for further exploitation such as prediction, recommendation etc. Table 6 shows the results.

We recall that, if we rely exclusively on a single data source such as the Crowdsourcing user (UserC) or the CCTV (ID1) the final information would have been incorrect since both haven't detect any congestion. From another side, the use of all the knowledge acquired from all data sources and the consideration of the meteorological conditions helped to find a more accurate and complete results.

The results of this mode are useful for performing a statistical and/or predictive analysis on the traffic state in order to suggest debottlenecking solutions.

*Streaming Mode*

To involve citizens in the debottlenecking of urban areas the system must be able to identify the state of traffic in a near-real-time. We propose a layer that analyzes streaming traffic data, detects congested roads in order to be used lately to recommend alternative route to citizens.

The streaming processing is thus based solely on the sources that provide data in real time. To validate our approach, we consider Twitter data source. However, this model is applicable to all other sources that provide data in real time. The streaming mode performs the following process: First Twitter Streaming API collects tweets that contain the appropriate keywords. Streaming data flew to our Kafka cluster, which transfers data every to spark streaming. The latest extracts metadata and analyzes tweets using machine Bayesian model previously built during batch processing to detect congestions. Finally it calculates the accuracy of the results based on the reliability index calculated during batch processing.

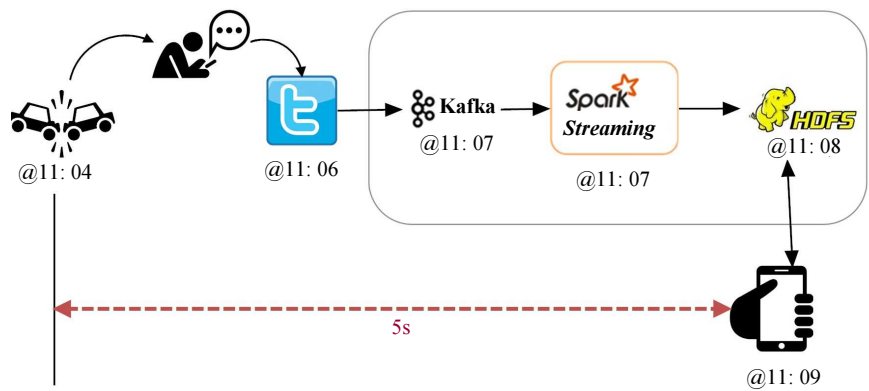

**Fig. 8:** Overall execution time of the streaming process

We have developed a solution to analyze twitter data and identify tweets talking about congestion and detects it location and time. Figure 8 shows the overall execution time of the streaming process which represents the time interval between the happenings of the event until its availability to the user. This time is equal to 5s which is an acceptable timeframe for this case study. However, considering that we did the evaluation in a single node machine, we believe that the performance of the system could be highly improved since all the used techniques and technologies support horizontal scalability.

*Querying Layer*

The Querying layer receives the departure and arrival coordinates of each user. Then the system identifies the different routes connecting these points using Google Map API and looks for the optimal route that avoids congestions. The latest are stored in the streaming views or could be predicted using the prebuilt models during the batch process. Therefore, the system identifies the roads sections of each route then looks for the states of each one in the streaming view. If no information found, the system predicts the road state. Finally the system displays the global insight with a percentage of reliability based on the reliability of the data source.

## Conclusion and Perspectives

We have presented in this paper an implementation and adaptation of the lambda architecture for multimodal Big Data integration. In the batch processing mode we have presented the different engines for processing different types of data in order to extract the knowledge that is merged later in order to provide global accurate insight. The stream processing mode analyzes incoming data in real-time or near-real-time. We have implemented our solution in the context of urban traffic management systems in order to help users access near-real-time information about traffic states by analyzing tweet streams. This implementation could clearly be enlarged to other data sources providing different real-time data types such as images and videos provided by CCTVS and aerial vehicles which constitute our ongoing work. In the midterm we are thinking about replacing the sources' reliability index computation by a new mechanism that identifies automatically the reputation of each data source. Another perspective aims to provide users with recommendations about alternative and less congested roads. Finally, we are thinking about evaluating our solution in a in a high performance cluster.

## Acknowledgment

The authors of this paper gratefully thank Miss. Boufarissi Fatima zahrae, graduate students from the School of Information Sciences in Rabat for their contribution to the validation of this work.

## Author's Contributions

**Siham Yousfi:** Accomplished all experiments, contributed to the design of the research plan and to the writing of the manuscript.

**Maryem Rhanoui:** Contributed to the design of the research plan and to the writing of the manuscript.

**Dalila Chiadmi:** Contributed to the design of the research plan and revised the manuscript.

## Ethics

This article is original and contains unpublished material. The corresponding author confirms that all of the other authors have read and approved the manuscript and there are no ethical issues involved.